\documentclass[letterpaper, 10 pt, conference]{ieeeconf}  

\IEEEoverridecommandlockouts                              

\usepackage{amsmath} 
\usepackage{amssymb}  
\usepackage{ascmac}
\usepackage{xcolor}
\usepackage{graphicx}
\usepackage{cite}
\usepackage{algorithm}
\usepackage{algorithmic}
\usepackage{multirow}
\usepackage{subcaption}
\usepackage{url}

\title{\LARGE \bf
OIPP: Object-Adaptive Impact Point Predictor \\ for Catching Diverse In-Flight Objects
}

\author{Ngoc Huy Nguyen$^{1}$, Kazuki Shibata$^{1}$ and Takamitsu Matsubara$^{1}$
\thanks{$^{1}$All the authors are with the Division of Information Science, Graduate School of Science and Technology, Nara Institute of Science and Technology (NAIST), Nara, Japan}
}

\begin{document}

\maketitle
\thispagestyle{empty}
\pagestyle{empty}

\begin{abstract}
In this study, we address the problem of in-flight object catching using a quadruped robot with a basket. Our objective is to accurately predict the impact point, defined as the object's landing position. This task poses two key challenges: the absence of public datasets capturing diverse objects under unsteady aerodynamics, which are essential for training reliable predictors; and the difficulty of accurate early-stage impact point prediction when trajectories appear similar across objects. To overcome these issues, we construct a real-world dataset of 8,000 trajectories from 20 objects, providing a foundation for advancing in-flight object catching under complex aerodynamics. We then propose the Object-Adaptive Impact Point Predictor (OIPP), consisting of two modules: (i) an Object-Adaptive Encoder (OAE) that extracts object-dependent representations from motion histories, and (ii) an Impact Point Predictor (IPP) that estimates the impact point from these representations. Two IPP variants are implemented: a Neural Acceleration Estimator (NAE)-based method that predicts trajectories and derives the impact point, and a Direct Point Estimator (DPE)-based method that directly outputs it. Experimental results show that our dataset is more diverse and complex than existing datasets, and that our method outperforms baselines on both 15 seen and 5 unseen objects. Furthermore, we show that improved early-stage prediction enhances catching success in simulation and demonstrate the effectiveness of our approach through real-robot experiments. The dataset and demonstration are available at \url{https://sites.google.com/view/robot-catching-2025}.
\end{abstract}

\section{Introduction}
Accurate prediction of the object’s future state is essential for robotic catching of in-flight objects~\cite{Kim-RAS2012, Kim-TRO2014, Lampariello-iros2011, Salehian-TRO2016, NAE-iros2021, Hu-iros2023}.
Owing to limited flight time, the robot must predict the object’s future state from only a short segment of its motion history in the early stage. In addition, the prediction must remain accurate under complex aerodynamic effects that make trajectories deviate significantly from simple parabolic motion.

\begin{figure}[t]
    \centering 
    \includegraphics[width=0.45\textwidth]{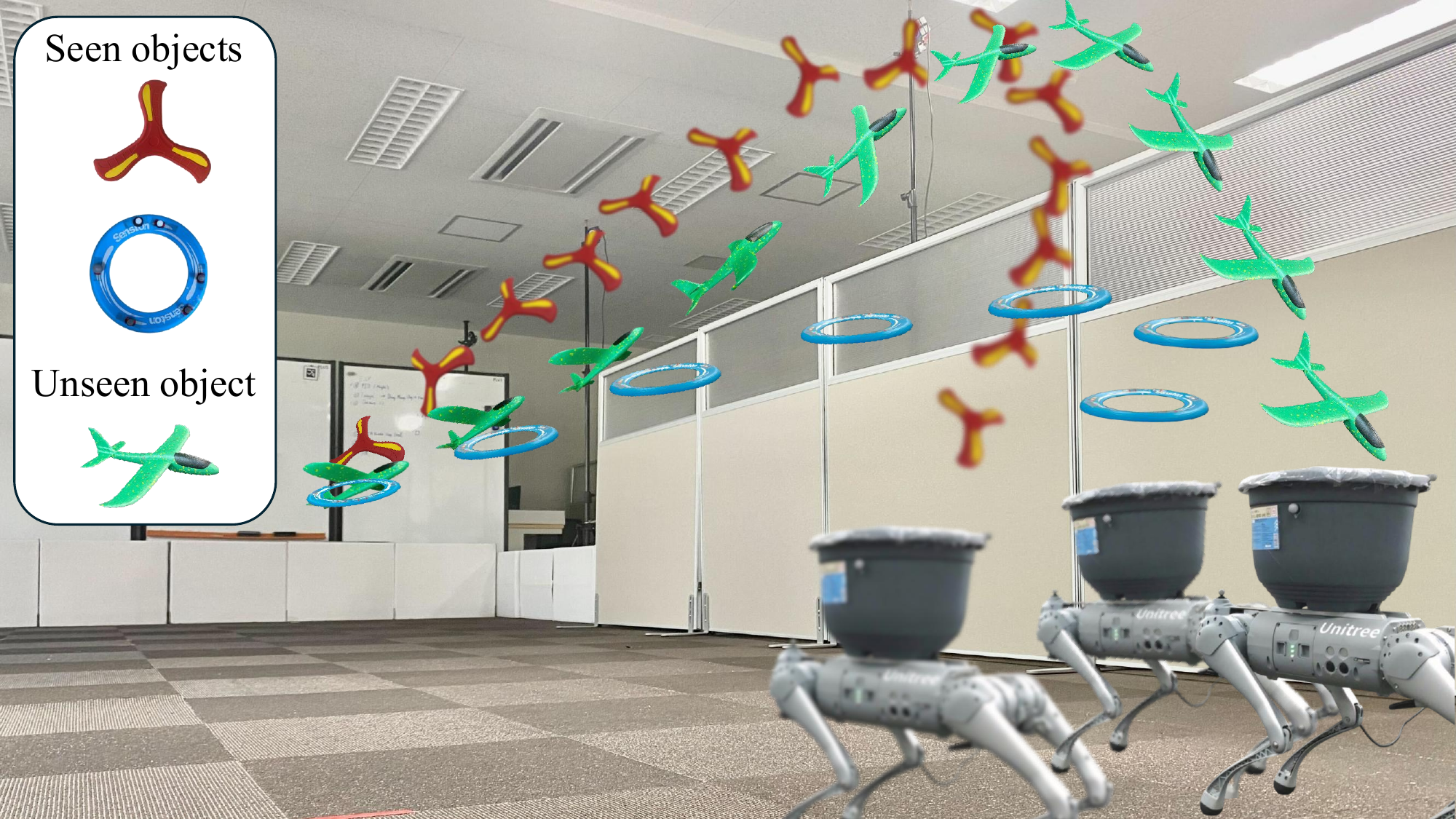}
    \caption{Catching diverse in-flight objects with complex aerodynamics using a quadruped robot}
    \label{fig:pull_figure}
\end{figure}

In this study, we address in-flight object catching using a quadruped robot equipped with a basket mounted on its body, as illustrated in Fig.~\ref{fig:pull_figure}. While manipulator-based catching~\cite{Kim-RAS2012, Kim-TRO2014, Lampariello-iros2011, Salehian-TRO2016, NAE-iros2021, Hu-iros2023} can intercept objects at various points along their trajectories, our task is constrained to a catching plane at a fixed height. Consequently, our objective is not to predict the entire trajectory but to accurately estimate the impact point \cite{Forrai-icra2023}, defined as the intersection between the object’s trajectory and the catching plane.

Predicting the impact point of diverse in-flight objects poses two key challenges.
First, there is no public dataset that captures complex aerodynamics across diverse objects, which is essential for training reliable predictors. Existing datasets are limited to six objects with mostly near-parabolic trajectories~\cite{NAE-iros2021}, while physics simulations diversify object shapes~\cite{Hu-iros2023} but fail to capture unsteady aerodynamic phenomena such as lift variation, Magnus forces, and vortex shedding.
Second, existing methods often fail to learn object-dependent representations from short motion histories, particularly in the early stage where trajectories appear similar across different objects. This makes it difficult to distinguish trajectories, leading to inaccurate predictions. Moreover, predicting trajectories of unseen objects is particularly challenging: even when the dynamics are similar to those observed during training, such similarities are hard to identify from short early-stage trajectories. Therefore, it is crucial to construct a dataset that captures complex aerodynamics and develop a prediction method that enables accurate early-stage impact point prediction for diverse in-flight objects.

In this study, we first construct a real-world dataset
that captures complex aerodynamic effects across diverse objects. 
The dataset comprises 2,000 measured trajectories from 20 objects (100 per object), which are further expanded to 8,000 trajectories through translational and rotational augmentation. 
This dataset provides the foundation for developing and evaluating our prediction method and contributes to advancing in-flight object catching under complex aerodynamics.

We then propose the Object-Adaptive Impact Point Predictor (OIPP) for catching diverse in-flight objects. OIPP consists of two key modules: (i) an Object-Adaptive Encoder (OAE) that extracts object-dependent representations from motion histories, and (ii) the impact point predictor (IPP), which predicts the impact point from these representations.
The OAE embeds historical states including position, velocity, and acceleration into a feature space where trajectories with similar dynamics are mapped close to each other.
This improves early-stage prediction accuracy and facilitates generalization to unseen objects by relating their trajectories to dynamically similar ones observed during training.
For the IPP module, we propose two approaches with complementary tradeoffs. The Neural Acceleration Estimator (NAE)-based method~\cite{NAE-iros2021} prioritizes accuracy and adaptability to varying catching heights by autoregressively rolling out the future trajectory and deriving the impact point, but incurs higher computational cost. In contrast, the Direct Point Estimator (DPE)-based method prioritizes faster impact point prediction by directly estimating the impact point from historical states, but yields lower accuracy and is restricted to fixed height catching.

To confirm the effectiveness of our approach, we first show that our dataset contains more diverse and complex trajectories than existing datasets~\cite{NAE-iros2021}. Using this dataset, our method outperforms baselines in impact point prediction accuracy for both 15 seen and 5 unseen objects. We further demonstrate that improved early-stage prediction increases the catching success rate in simulation. Finally, we demonstrate the effectiveness of our approach through real-world catching experiments with a quadruped robot.

The main contributions of this study are summarized as follows:
\begin{itemize}
    \item We construct a novel real-world dataset of 8,000 trajectories from 20 diverse objects, capturing complex aerodynamics not presented in existing datasets.
    \item We propose OIPP, a framework for impact point prediction of diverse in-flight objects.
    \item We conduct ablation studies comparing LSTM and Transformer encoders for the OAE and show improved prediction accuracy on both seen and unseen objects.
    \item We show that enhanced early-stage prediction increases catching success rates in simulation and real-robot demonstrations.
\end{itemize}

\section{Related work}
\subsection{Physics-based Approaches}
Existing approaches \cite{Ritz-iros2012, Kober-humanoids2012, Lippiello-icra2012, Wang-ral2022, Abeyruwan-L4D2023, Forrai-icra2023, You-iros2023, Zhao-iros2023, Schakkal-iros2024, Yan-TRO2024, Tassi-IJRR2025} predict the future states of flying objects based on Newtonian dynamics, modeling them as point masses using simplified aerodynamic models. To improve prediction accuracy under observation noise, various Kalman filter variants have been introduced, including the Extended Kalman Filter (EKF)\cite{Huang-iros2021,Yu-TMECH2023} and discrete-time filters\cite{Müller-iros2011,Dong-iros2020}. Furthermore, Jia et al.~\cite{Jia-TRO2024} employed Koopman operator to estimate complex aerodynamic disturbances for accurate prediction.

While these approaches are computationally efficient, they require accurate modeling of object-specific dynamics, which becomes impractical for objects with complex aerodynamics. In contrast, our method employs a deep learning approach without relying on dynamics modeling, thereby enabling application to a wider range of objects.

\subsection{Regression-based Approaches}
Prior works have adopted regression-based methods such as Support Vector Regression (SVR)\cite{Lampariello-iros2011, Kim-RAS2012,Kim-TRO2014} and Gaussian Process Regression (GPR)~\cite{Salehian-TRO2016} to predict flying object motion based on time-series trajectory data.

While these methods can be applied without relying on dynamics models, they often fail to achieve accurate predictions when dealing with objects exhibiting complex aerodynamics, as demonstrated in~\cite{NAE-iros2021}. In contrast, our study adopts a deep learning approach that models temporal patterns in historical motion, leading to improved prediction accuracy for objects with complex aerodynamics.

\subsection{Deep Learning-based Approaches}
Advances in deep learning have significantly improved trajectory prediction for in-flight objects. LSTM-based models~\cite{NAE-iros2021, Hu-iros2023, Monforte-cvpr2023} predict trajectories from time-series data, while Transformer-based models~\cite{Lee-access2024} have been explored for related motion prediction tasks such as rolling objects on planar surfaces from RGB images. However, existing real-world datasets remain limited—for example, prior work considers only six objects with mostly near-parabolic trajectories~\cite{NAE-iros2021}—and simulation-based datasets, although they diversify shapes~\cite{Hu-iros2023}, do not capture unsteady aerodynamics observed in real flights. Moreover, these approaches struggle with diverse objects because early-stage trajectories often appear similar, making it difficult to extract object-dependent representations, especially for unseen objects.

In contrast, our study addresses these limitations by constructing a diverse real-world dataset that captures complex aerodynamics, providing a foundation for advancing in-flight object catching. We further propose OIPP, which learns object-dependent representations that capture dynamics differences from short motion histories and integrates them with both an NAE-based predictor and a DPE-based predictor, enabling accurate impact point prediction for diverse objects under complex aerodynamics.

\section{Method}
\begin{figure*}[t!]
    \centering

    \begin{subfigure}[b]{0.95\linewidth}
        \centering
        \includegraphics[width=\linewidth]{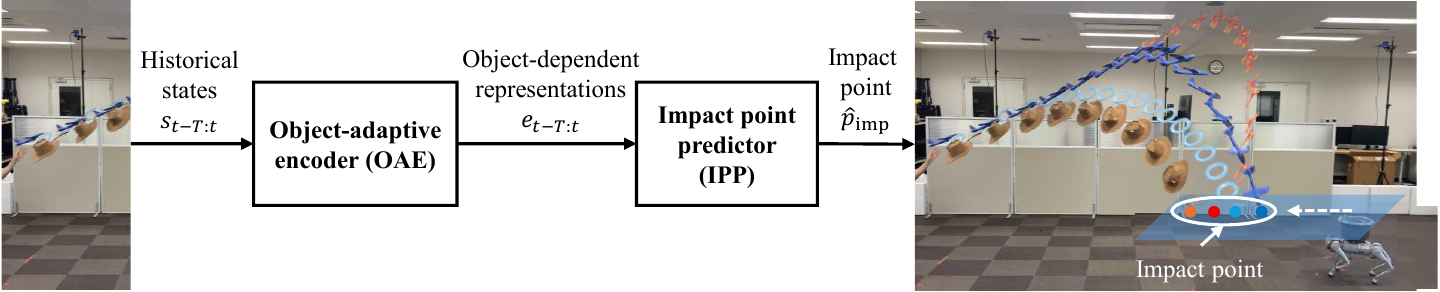}
        \caption{Overview}
        \label{fig:overview}
    \end{subfigure}

    \vspace{0.5em}  

    \begin{subfigure}[b]{0.39\linewidth}
        \centering
        \includegraphics[width=\linewidth]{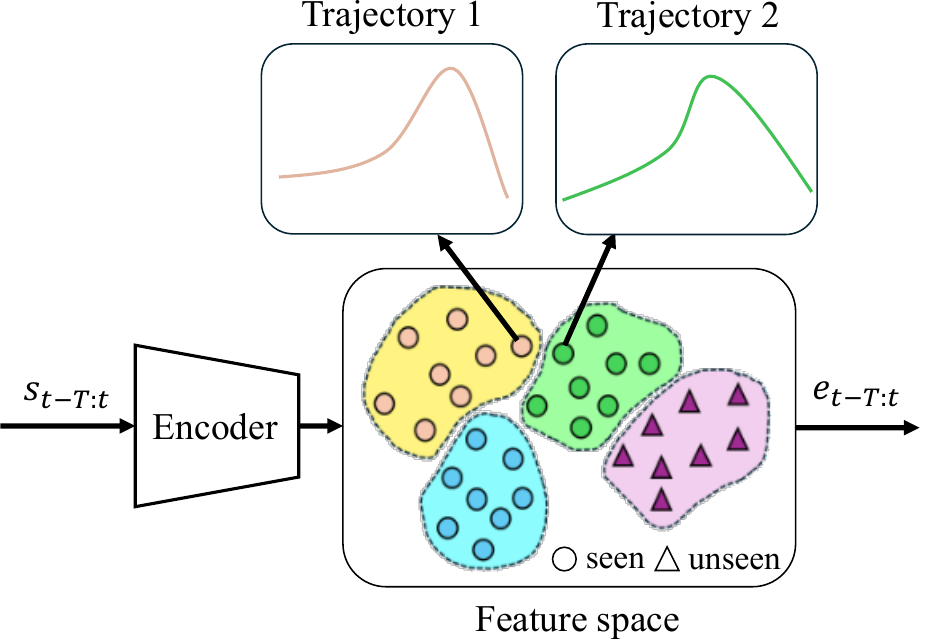}
        \caption{OAE}
        \label{fig:DIPP}
    \end{subfigure}
    \hfill
    \begin{subfigure}[b]{0.56\linewidth}
        \centering
        \includegraphics[width=\linewidth]{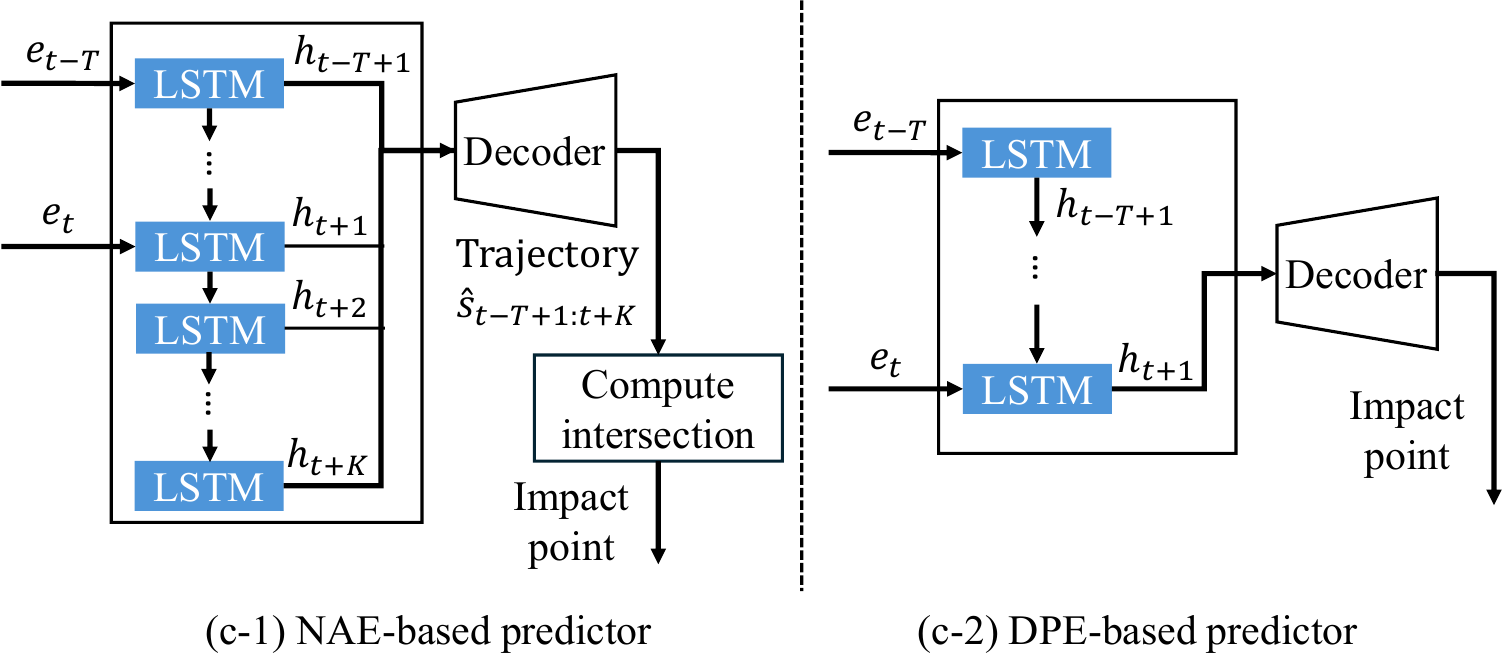}
        \caption{IPP}
        \label{fig:IPP}
    \end{subfigure}

    \caption{OIPP framework for catching diverse in-flight objects}
    \label{fig:framework}
\end{figure*}

This section introduces OIPP for catching diverse in-flight objects, as shown in Fig.~\ref{fig:framework}. OIPP consists of two modules: the OAE and IPP modules.

The OAE maps historical motion states into a feature space where trajectories with similar dynamics are mapped close together and dissimilar ones are mapped farther apart, enabling object-dependent representation learning and generalization to unseen objects. The IPP estimates the impact point from these representations. We consider two variants of the IPP: one based on the NAE method, which learns dynamics to predict trajectories and derives the impact point, and a DPE-based method, which directly outputs the impact point from historical states. 

To further improve prediction accuracy, we introduce the Impact Point Enhanced (IPE) loss, which explicitly minimizes the gap between predicted and actual impact points. The OAE and IPP are jointly optimized with the IPE loss, yielding accurate impact point predictions and improved catching performance across diverse objects.

\subsection{OAE}
Existing approaches~\cite{NAE-iros2021, Hu-iros2023} learn to predict future trajectories from motion histories, but generalization across objects remains difficult because early-stage trajectories often appear similar. In other domains such as object transport, representation learning has been shown to improve generalization across diverse objects~\cite{Jeon2024}. Building on this insight, we employ an OAE that learns object-dependent representations from motion histories.

Given the input sequences $\boldsymbol{s}_{t-T:t}$, the encoder $E(\cdot)$ in the OAE computes object-dependent representations as
\begin{equation}
\boldsymbol{e}_{t-T:t} = E(\boldsymbol{s}_{t-T:t}),
\label{eq:encoder}
\end{equation}
where $\boldsymbol{s}_t = [\boldsymbol{x}_t^\top, \boldsymbol{v}_t^\top, \boldsymbol{a}_t^\top]^\top \in \mathbb{R}^9$ represents the object state at time $t$, composed of position $\boldsymbol{x}_t \in \mathbb{R}^3$, velocity $\boldsymbol{v}_t \in \mathbb{R}^3$, and acceleration $\boldsymbol{a}_t \in \mathbb{R}^3$. By learning object-dependent representations from historical motion data, the OAE can improve trajectory prediction across diverse objects.

\subsection{IPP}
While existing studies~\cite{NAE-iros2021, Hu-iros2023} focus on trajectory prediction, our objective is impact point prediction. To this end, we propose two approaches: an NAE-based predictor that learns a dynamics model to predict future trajectories and derives the impact point, and a DPE-based predictor that directly estimates the impact point from historical states.

\textbf{NAE-based Predictor:}
Given the object-dependent representations, the hidden state is recursively updated using
\begin{equation}
\boldsymbol{h}_{t-T+i+1} = 
\begin{cases} 
\mathrm{LSTM}(\boldsymbol{e}_{t-T+i}, \boldsymbol{h}_{t-T+i}) & \text{if } 0 \leq i \leq T \\
\mathrm{LSTM}(\boldsymbol{h}_{t-T+i}) & \text{if } T < i \leq T + K
\end{cases}
\label{eq:autoregression}
\end{equation}
where the representations $\boldsymbol{e}_{t-T+i}$ are used for $0 \leq i \leq T$, and the hidden state $\boldsymbol{h}_{t-T+i}$ is used for $i > T$, since future representations are unavailable. Here, $K$ denotes the number of steps required to reach a plane $\mathcal{P}$ at a fixed height, as shown in Fig.~\ref{fig:overview}.

At each step $i$, the hidden state is passed to a decoder network $D(\cdot)$ to generate the predicted future state:
\begin{equation}
\boldsymbol{\hat{s}}_{t-T+i+1} = D_{\mathrm{NAE}}(\boldsymbol{h}_{t-T+i+1})
\label{eq:decoder}
\end{equation}
Finally, the impact point $\hat{\boldsymbol{p}}_{\mathrm{imp}} \in \mathbb{R}^3$ is obtained as the intersection of the predicted trajectory with the plane $\mathcal{P}$.

\textbf{DPE-based Predictor:}  
Unlike the NAE-based predictor, which generates the future trajectory and derives the impact point, the DPE-based predictor directly outputs the impact point. Specifically, instead of autoregressively applying \eqref{eq:autoregression} and \eqref{eq:decoder} for future steps, it only computes
\begin{equation}
\hat{\boldsymbol{p}}_{\mathrm{imp}}
= D_{\mathrm{DPE}}(\boldsymbol{h}_{t+1}),
\end{equation}
where $\hat{\boldsymbol{p}}_{\mathrm{imp}}$ is obtained directly from the hidden state.

While the DPE-based predictor is computationally efficient, it is restricted to fixed-height catching. In contrast, the NAE-based predictor autoregressively predicts future trajectories as in Eq.~\eqref{eq:autoregression}, which leads to higher computational cost but enables applicability beyond fixed-height catching.

\subsection{Training Objective}
Unlike existing trajectory prediction methods~\cite{NAE-iros2021, Hu-iros2023}, we introduce the IPE loss, which explicitly penalizes errors at the impact point. This loss directly reduces the gap between predicted and true impact points, enabling more accurate estimation. The training objectives are then defined according to the type of predictor.

\subsubsection{NAE-based Predictor}
Since this predictor generates the entire future trajectory, its training objective combines (i) teacher-forcing and reconstruction losses~\cite{NAE-iros2021}, and (ii) the proposed IPE loss, given by
\begin{align}
\mathcal{L}_\text{NAE} 
&= \underbrace{\frac{1}{T+1} \sum_{i=1}^{T+1} \left\| \boldsymbol{s}_{t-T+i} - \hat{\boldsymbol{s}}_{t-T+i} \right\|^2}_{\text{teacher forcing}} \nonumber \\
&+ \underbrace{\frac{1}{T+1} \sum_{i=0}^{T} \left\| \boldsymbol{s}_{t-T+i} - D_{\mathrm{NAE}}\left(E\left(\boldsymbol{s}_{t-T+i}\right)\right) \right\|^2}_{\text{reconstruction}} \nonumber \\
&+ \underbrace{\frac{1}{K-1}\sum_{i=2}^{K} \left\| \boldsymbol{s}_{t+i} - \hat{\boldsymbol{s}}_{t+i} \right\|^2}_{\text{trajectory alignment (IPE)}} 
+ \underbrace{\left\| \boldsymbol{p}_{\mathrm{imp}} - \hat{\boldsymbol{p}}_{\mathrm{imp}} \right\|^2}_{\text{impact point (IPE)}}.
\end{align}
The first two terms ensure that the encoder and decoder preserve sufficient information in the representations, while the last two terms constitute the proposed IPE loss, which aligns the predicted trajectory with the ground truth up to the impact point and explicitly penalizes the impact point error.

\subsubsection{DPE-based Predictor}
This predictor does not generate trajectories; instead, it directly estimates the impact point from historical states. Accordingly, its training objective consists of the reconstruction loss and the IPE loss:
\begin{align}
\mathcal{L}_\text{DPE}
&= \frac{1}{T+1} \sum_{i=0}^{T} 
\left\| \boldsymbol{s}_{t-T+i} - D_\text{DPE}\left(E\left(\boldsymbol{s}_{t-T+i}\right)\right) \right\|^2
\nonumber \\
&+ \left\| \boldsymbol{p}_{\mathrm{imp}} - \hat{\boldsymbol{p}}_{\mathrm{imp}} \right\|^2.
\label{eq:loss_reg}
\end{align}
The reconstruction loss ensures that the encoder preserves sufficient information in the latent representations, while the IPE loss directly evaluates the discrepancy between the predicted and actual impact points.

\section{Experiment}
In this section, we validate the effectiveness of the proposed method through experiments using a real-world trajectory dataset of diverse in-flight objects. The key research questions in this experiment are as follows:

\begin{itemize}
    \item \textbf{RQ1:} Does our dataset include more diverse and complex trajectories than existing datasets?
    \item \textbf{RQ2:} Can the proposed method improve early-stage prediction for multiple seen and unseen objects?
    \item \textbf{RQ3:} Can the better early-stage prediction improve the catching performance?
    \item \textbf{RQ4:} Can the proposed method be successfully demonstrated in the real world?
\end{itemize}

In the following experiments, training and quantitative evaluation are conducted in simulation, while real-world experiments are used for demonstration, because achieving identical initial conditions in physical throwing is difficult, making fair quantitative comparison challenging.

\subsection{Dataset}
\label{sec:Dataset}

\begin{figure}[t]
  \centering
  \includegraphics[width=0.85\linewidth]{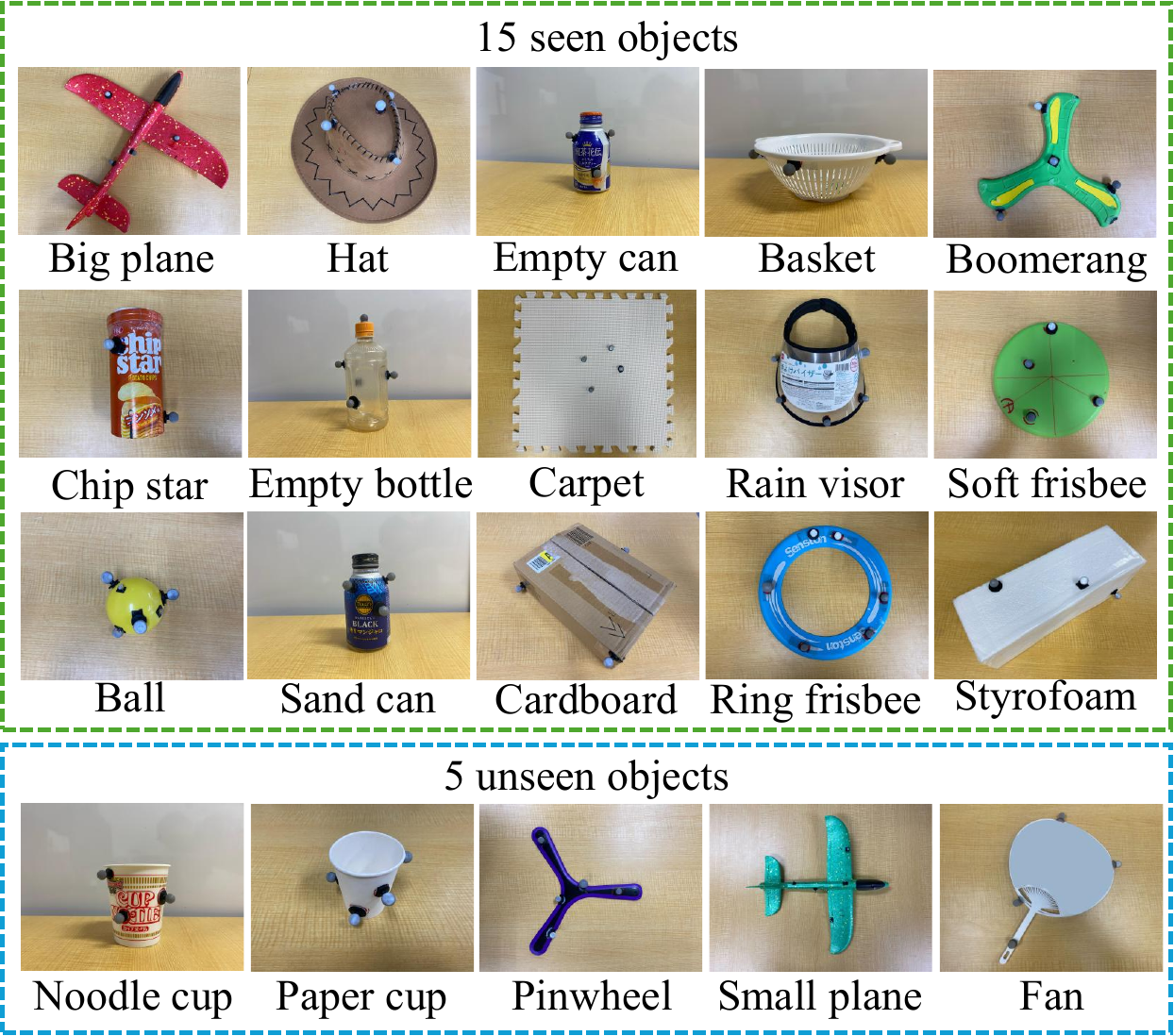}
  \caption{20 objects used for experiment}
  \label{fig:dataset}
\end{figure}

We constructed a dataset of 20 flying objects, as shown in Fig.~\ref{fig:dataset}. Each object was hand-thrown and recorded at 120 Hz using a motion capture system, yielding 100 trajectories per object. The data were augmented to 400 trajectories per object with random horizontal translations of up to $\pm 0.3\mathrm{m}$ in the $x$–$z$ plane and random rotations of up to $\pm 10^\circ$ about the vertical $y$-axis around the initial position. The trajectories covered distances of $2.9$--$4.2\,\mathrm{m}$ and maximum heights of $2.5$--$2.8\,\mathrm{m}$, keeping full flights within the motion-capture range.

\subsection{Compared Methods}
To evaluate the effectiveness of the proposed method, comparisons were made with the following baseline approaches:
\begin{itemize}
    \item \textbf{Newton}~\cite{Forrai-icra2023}: a Newtonian mechanics-based method that applies RANSAC for outlier removal and then fits a parabolic trajectory using least squares.
    \item \textbf{SVR}~\cite{Kim-RAS2012}: an SVR-based regression method using historical position data.
    \item \textbf{NAE}~\cite{NAE-iros2021}: an LSTM-based trajectory prediction method, which serves as the basis for our NAE-based predictor.
\end{itemize}

Furthermore, we evaluate the following variants of our proposed method:
\begin{itemize}
    \item \textbf{DPE}: a direct impact point predictor without OAE, used to assess the effect of representation learning.
    \item \textbf{OIPP-NAE}: 
    a variant that combines the OAE with an NAE-based predictor and differs from the NAE baseline by using OAE-based representations and the proposed IPE loss with impact-point-aligned trajectories rather than a fixed prediction horizon.

    \item \textbf{OIPP-DPE}: a variant that combines the OAE with a DPE-based predictor.
\end{itemize}

\subsection{Training Setup}
We implemented \textbf{NAE}, \textbf{DPE}, \textbf{OIPP-NAE}, and \textbf{OIPP-DPE} using PyTorch. Each model is optimized using the Adam optimizer with a batch size of 512 and trained for up to $3.0 \times 10^{4}$ epochs. The learning rate is set to $3.0 \times 10^{-5}$ for \textbf{DPE}, \textbf{OIPP-NAE}, and \textbf{OIPP-DPE}, and $1.0 \times 10^{-4}$ for \textbf{NAE}. These parameters are determined through empirical tuning based on validation performance.

We examined multiple encoder choices for the OAE, including FC, LSTM, and Transformer, and adopted the LSTM encoder based on the ablation study in Appendix~\ref{sec:ablation_oae_encoder}. \textbf{OIPP-NAE} and \textbf{OIPP-DPE} share the same architecture, consisting of a single-layer LSTM encoder, followed by a two-layer LSTM module and a single fully connected (FC) decoder. \textbf{OIPP-NAE} outputs the future trajectory, while \textbf{OIPP-DPE} outputs only the impact point.
The \textbf{DPE} is constructed by replacing the LSTM encoder in \textbf{OIPP-NAE} with a single FC layer, while keeping the rest of the network identical. All hidden layers contain 128 units. The architecture of \textbf{NAE} is implemented following~\cite{NAE-iros2021}.

Training is conducted on a dataset consisting of 15 out of 20 flying objects, while the remaining 5 objects are used to evaluate the generalization to unseen objects. For each of the 15 training objects, the dataset is divided into 80\% for training, 10\% for validation, and 10\% for testing.

\subsection{Evaluation Metrics}
To quantitatively evaluate the proposed model and baselines, we employ the following two metrics:

\textbf{Impact Point Error (IE):}
IE is defined as the Euclidean distance between the predicted and ground-truth impact points.

\textbf{Success Rate (SR):}
SR is defined as the ratio of test trials in which the ground-truth impact point falls within the radius of a circular basket, which is mounted at a predefined height on the robot.

\subsection{RQ1: Does Our Dataset Include More Diverse and Complex Trajectories than Existing Datasets?}
\label{sec:RQ1-data-analysis}

\subsubsection{Setup}
To assess the complexity and diversity of object trajectories, we analyzed the real-world trajectories included in our dataset and the NAE dataset~\cite{NAE-iros2021}. For quantitative analysis, we defined the Parabola Deviation Score (PDS) as the difference between each trajectory and its approximated parabolic curve. A higher PDS value indicates a significant deviation from a parabolic motion, suggesting that the trajectory exhibits complex motion patterns that deviate from simple Newtonian dynamics. See Appendix~A for details of the PDS computation.
 
\subsubsection{Results}
\begin{figure}[t]
  \centering
  \includegraphics[width=1.0\linewidth]{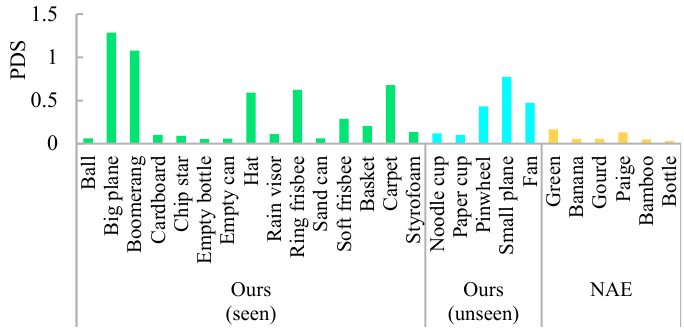}
  \caption{Dataset analysis for our dataset and the NAE dataset}
  \label{fig:dataset_analysis}
\end{figure}

The PDS for individual objects in each dataset is shown in Fig.~\ref{fig:dataset_analysis}. The NAE dataset exhibits low PDS values for six objects, indicating that their trajectories closely follow parabolic motion. In contrast, our dataset demonstrates higher PDS values for both seen and unseen objects.

In summary, our dataset presents more diverse and complex trajectories compared to existing datasets.

\subsection{RQ2: Can the Proposed Method Improve Early-stage Prediction for Multiple Seen and Unseen Objects?}
In this experiment, we evaluate the prediction performance of each method using real-world trajectories for multiple seen and unseen objects.

\begin{figure}[t!]
    \centering

    \begin{subfigure}[b]{0.95\linewidth}
        \centering
        \includegraphics[width=\linewidth]{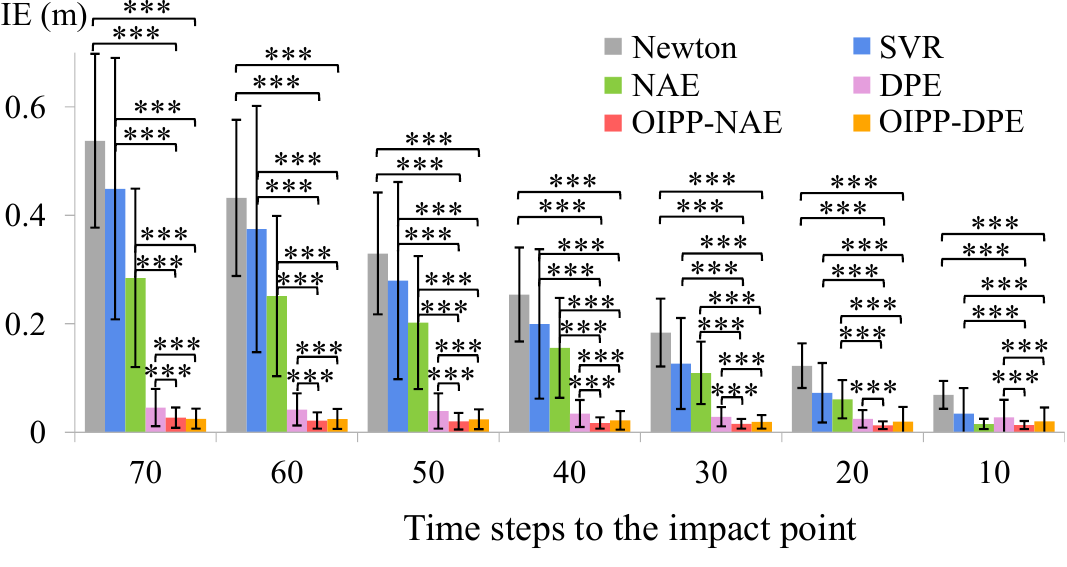}
        \caption{15 seen objects}
        \label{fig:ie_seen}
    \end{subfigure}

    \vspace{0.5em}  

    \begin{subfigure}[b]{0.95\linewidth}
        \centering
        \includegraphics[width=\linewidth]{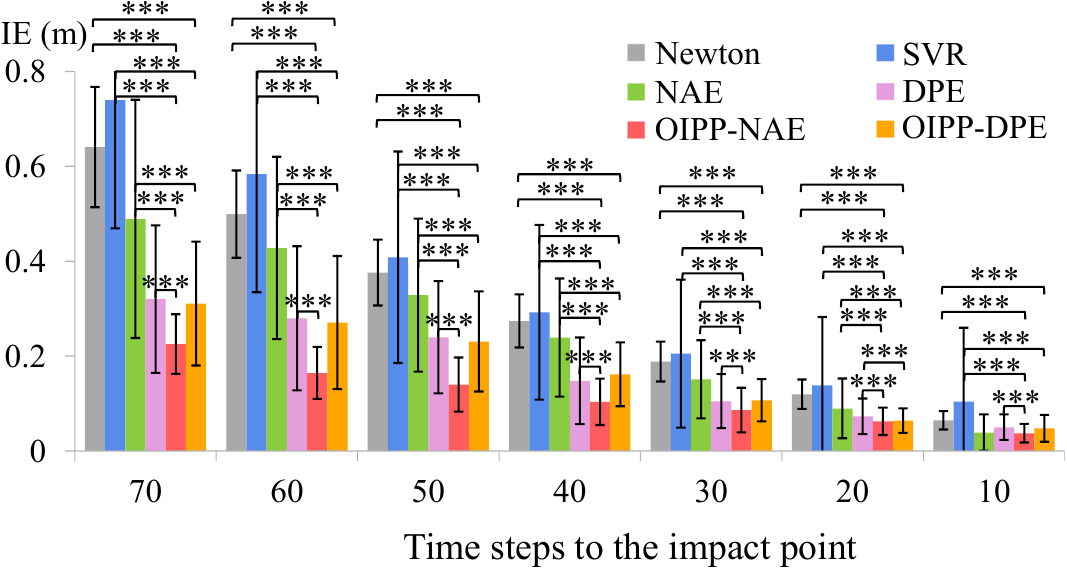}
        \caption{5 unseen objects}
        \label{fig:ie_unseen}
    \end{subfigure}

    \caption{Comparison of impact point errors. Statistically significant differences ($p < 0.05$) are denoted by \textsuperscript{***}. One timestep corresponds to 1/120 s.}
    \label{Impact_point_prediction_error}
\end{figure}

The IE values across different time steps to the impact point for each method are presented in Fig.~\ref{Impact_point_prediction_error}. \textbf{OIPP-NAE} and \textbf{OIPP-DPE} yield lower errors than the baselines and \textbf{DPE} for both seen and unseen objects. Notably, the performance gaps are larger when predicting from earlier observations, indicating that the proposed method improves early-stage impact point prediction. 

\begin{figure}[t!]
    \centering

    \begin{minipage}[t]{0.5\textwidth}
        \centering
        \includegraphics[width=\textwidth]{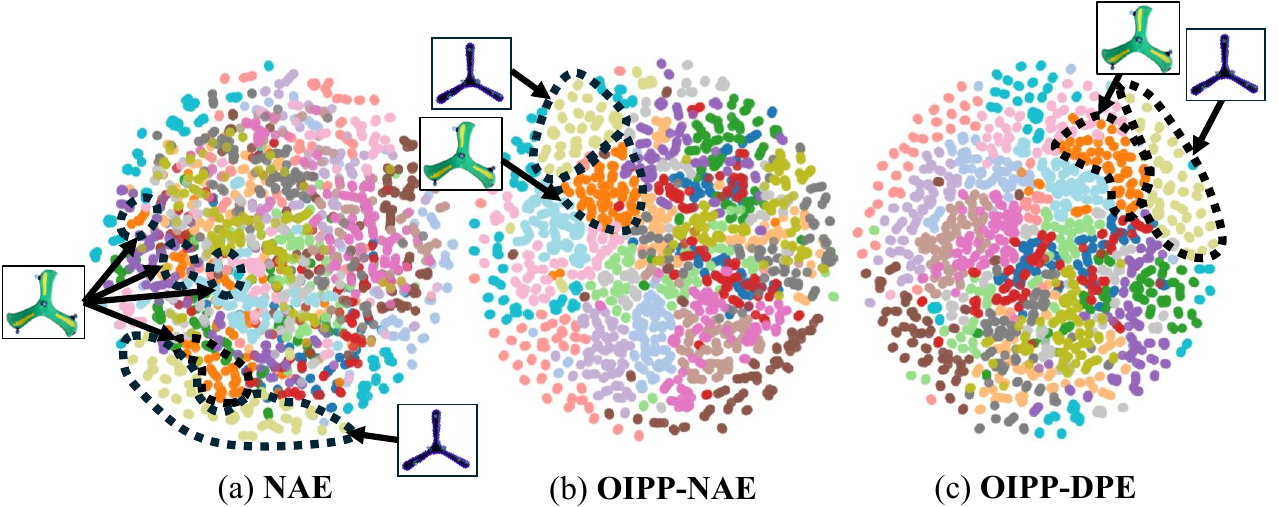}
    \end{minipage}

    \vspace{0.2em}
    \begin{minipage}[t]{0.5\textwidth}
        \centering
        \includegraphics[width=\textwidth]{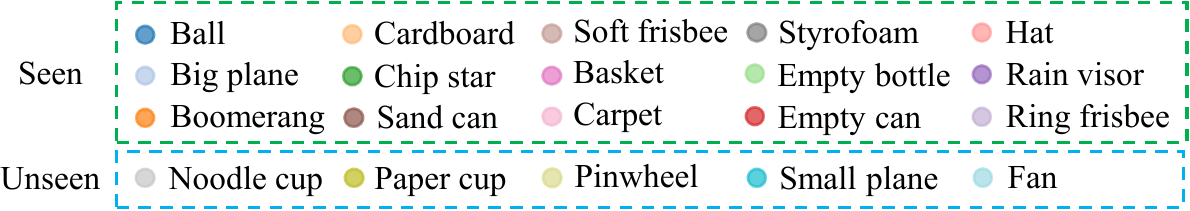}
    \end{minipage}

    \caption{Visualization of embedded features using t-SNE}
    \label{t_SNE_feature_spaces_of_methods_early_stage}
\end{figure}

To analyze the learned representations from early-stage trajectories, we visualize 2D t-SNE projections of the OAE representations, as shown in Fig.~\ref{t_SNE_feature_spaces_of_methods_early_stage}. The first five consecutive segments of each trajectory are defined as the early stage. The results show that \textbf{OIPP-NAE} and \textbf{OIPP-DPE} form a more structured organization of trajectories than \textbf{NAE}, where trajectories with similar dynamics tend to appear close to each other. For example, trajectories of the pinwheel appear near those of the boomerang, which share similar dynamics. Although \textbf{NAE} also places these two objects relatively close, the boomerang embeddings are more scattered, and several samples lie farther away from the pinwheel cluster. This suggests that our proposed method learns representations that capture such similarities between seen and unseen objects.

\begin{figure}[t!]
    \centering
    \begin{minipage}[t]{0.45\textwidth}
        \centering
        \includegraphics[width=\textwidth]{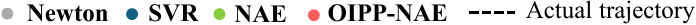}
    \end{minipage}
    \begin{subfigure}[b]{0.23\textwidth}
        \centering
        \vspace{0.5em}
        \includegraphics[width=\linewidth]{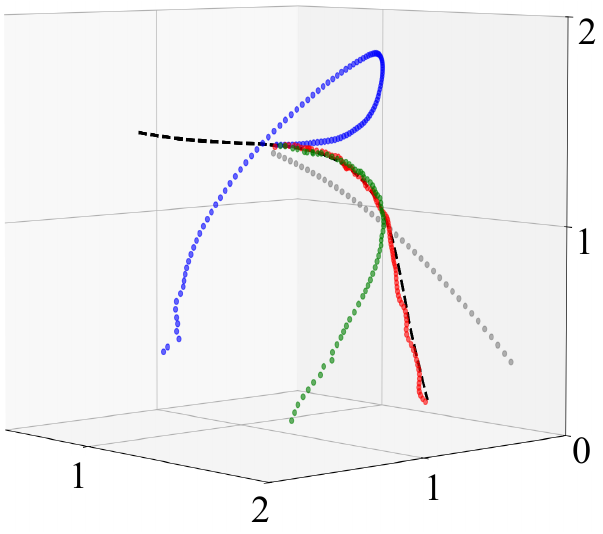}
        \caption{Big plane (seen)}
        \label{fig:trajectory_big_plane}
    \end{subfigure}
    \hfill
    \begin{subfigure}[b]{0.23\textwidth}
        \centering
        \includegraphics[width=\linewidth]{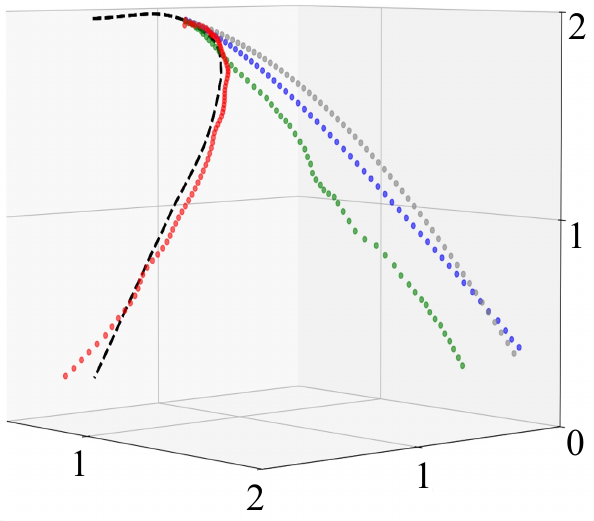}
        \caption{Fan (unseen)}
        \label{fig:trajectory_fan}
    \end{subfigure}

    \caption{A comparison of the predicted trajectories.}
    \label{fig:trajectory}
\end{figure}

Finally, a comparison of the predicted trajectories for the boomerang and pinwheel is shown in Fig.~\ref{fig:trajectory}. The results show that \textbf{OIPP-NAE} predicts trajectories that are closer to the actual trajectory for both objects compared to the other baseline methods.

Overall, the proposed method learns object-dependent representations from early-stage motion histories, resulting in improved prediction for multiple seen and unseen objects.

\subsection{RQ3: Can Better Early-stage Prediction Improve Catching Performance?}

\subsubsection{Setup}
his experiment evaluates whether improved prediction enhances catching under robot motion constraints. Since the robot cannot be instantaneously placed at the predicted impact point, catching success depends on how impact point estimates updated during flight guide robot motion. We conduct a simulation using our real-world trajectory dataset to evaluate catching success rate. The robot is modeled as a PID-controlled single integrator with a maximum velocity of 2.5~m/s, and its initial position is randomly sampled within a 0.3~m radius circle centered at the ground-truth impact point.

\begin{figure*}[t!]
    \centering
    \begin{subfigure}[b]{0.9\textwidth}
        \centering
        \includegraphics[width=\linewidth]{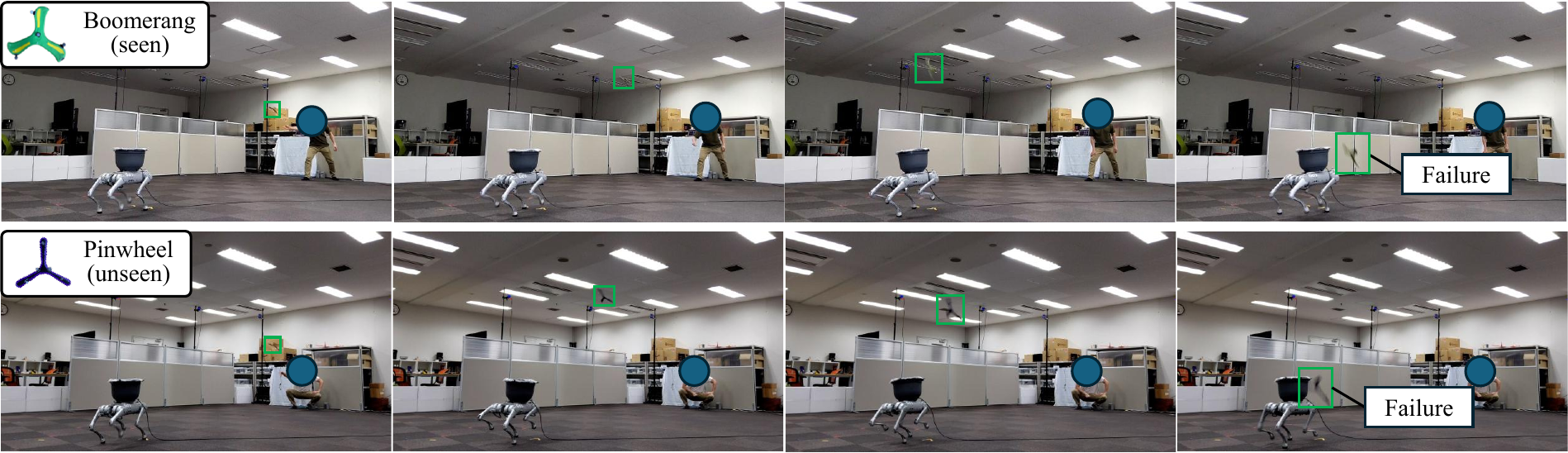}
        \caption{\textbf{NAE}}
        \label{fig:demo_NAE}
    \end{subfigure}
    \\[0.2cm]
    \begin{subfigure}[b]{0.9\textwidth}
        \centering
        \includegraphics[width=\linewidth]{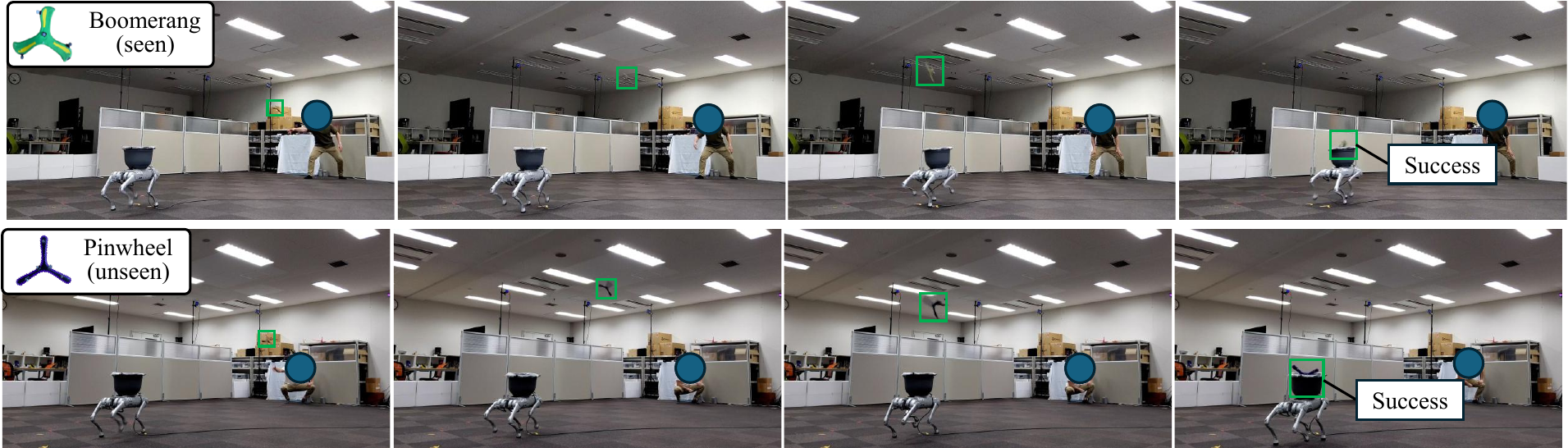}
        \caption{\textbf{OIPP-NAE}}
        \label{fig:demo_OIPP}
    \end{subfigure}
    \caption{Comparisons of real-robot demonstrations.}
    \label{fig:demo}
\end{figure*}

\subsubsection{Results}
Table~\ref{tab:catching_success} presents the catching success rates for different basket radii. The evaluation was conducted with 100 trials for each of 5 seen and 5 unseen objects. The results show that \textbf{OIPP-NAE} achieves the highest success rates for both seen and unseen objects across various basket radii. Notably, \textbf{OIPP-NAE} improves catching performance on unseen objects by producing more accurate early-stage impact point predictions based on motion histories.

Overall, improved early-stage prediction accuracy contributes significantly to the catching performance.

\begin{table}[t]
    \centering
    \small
    \caption{Catching success rates for different basket radii $r$ using 5 seen and 5 unseen objects. Each value represents the success rate for seen / unseen objects.}
    \begin{tabular}{lcccc}
    \hline
    Method   & $r=0.05$ & $r=0.10$ & $r=0.15$ & $r=0.20$ \\
    \hline
        \textbf{Newton}~\cite{Forrai-icra2023}     & 0.02/0.04 & 0.06/0.16 & 0.11/0.32 & 0.25/0.44 \\
        \textbf{SVR}~\cite{Kim-RAS2012}            & 0.06/0.06 & 0.17/0.17 & 0.35/0.32 & 0.52/0.47 \\
        \textbf{NAE}~\cite{NAE-iros2021}           & 0.11/0.08 & 0.36/0.26 & 0.66/0.46 & 0.83/0.66 \\
        \textbf{DPE}                               & 0.51/0.13 & 0.79/0.33 & 0.92/0.55 & 0.97/0.72 \\
        \textbf{OIPP-NAE}                          & 0.51/\textbf{0.16} & \textbf{0.84}/\textbf{0.38} & \textbf{0.96}/\textbf{0.59} & \textbf{0.98}/\textbf{0.78} \\
        \textbf{OIPP-DPE}                          & \textbf{0.54}/\textbf{0.16} & 0.80/0.35 & 0.94/0.55 & \textbf{0.98}/0.71 \\
    \hline
    \end{tabular}
    \label{tab:catching_success}
\end{table}

\subsection{RQ4: Can the Proposed Method be Successfully Demonstrated in the Real World?}

\subsubsection{Setup}
To demonstrate the practical applicability of our approach, we conducted real-world catching experiments using a quadruped robot. The robot position was controlled by a PID controller, with gains tuned to reach the predicted impact point as quickly as possible. Since \textbf{OIPP-NAE} achieved a higher SR than \textbf{OIPP-DPE}, we report real-world demonstrations only for \textbf{OIPP-NAE}. We evaluated two seen and two unseen objects, comparing \textbf{OIPP-NAE} with \textbf{NAE} under comparable conditions by keeping the robot’s initial pose, throwing motion, and target impact point as consistent as possible. The basket radius was 0.15~m.

\subsubsection{Results}
Figure~\ref{fig:demo} shows representative snapshots of the real-robot catching demonstrations for one seen object (boomerang) and one unseen object (pinwheel) using both \textbf{NAE} and \textbf{OIPP-NAE}. While the robot failed to catch these objects with \textbf{NAE}, \textbf{OIPP-NAE} successfully caught them. See the attached video for additional demonstrations.

In summary, the proposed method enables successful catching in real-robot experiments.

\section{Discussion}
This section discusses the limitations of the proposed method and outlines potential directions for improvement.

Our method assumes a basket mounted horizontally on the robot base, constraining the catching space to a fixed height. This limits applicability to platforms such as mobile manipulators that catch objects at varying poses. To address this, we combine our method with a catching pose quality network \cite{Hu-iros2023} that learns to score object poses for catching, and select the highest-scoring pose as the impact point.

The current framework relies solely on object trajectories. Incorporating human motion as an additional input for object identification is a promising direction, and future work will investigate its effect on prediction accuracy.

\section{Conclusion}

In this paper, we constructed a real-world dataset of 8,000 trajectories from 20 objects exhibiting complex aerodynamics and proposed OIPP for catching diverse in-flight objects. OIPP consists of an OAE that learns object-dependent representations from motion histories and an IPP that estimates the impact point with two
variants: a NAE-based method and a DPE-based method. 
Experiments showed that our dataset captures more diverse and complex trajectories than existing ones, and that OIPP outperforms baselines on seen and unseen objects. 
Furthermore, improved early-stage impact point prediction also enhanced catching success in simulation, while real-world experiments with a quadruped robot demonstrated the practical effectiveness of our approach.

As future work, we aim to validate the proposed method in settings with different catching mechanisms and robot platforms, such as quadrotors and mobile manipulators.

\appendix
\subsection{Evaluation Metrics for Trajectory Analysis}
This appendix describes the computation of the Parabola Deviation Score (PDS) used in the trajectory analysis.  
First, the parameters of the parabolic motion that best approximate each object trajectory are obtained as follows:
\begin{align}
\boldsymbol{v}_{0} &= \arg\min_{\boldsymbol{v}_{0}} \sum_{i=0}^{N_p} 
\left\| \boldsymbol{p}_i - \left(\boldsymbol{p}_0 + \boldsymbol{v}_0 t_i + \tfrac{1}{2} \boldsymbol{g} t_i^2 \right) \right\|^2 ,
\end{align}
where \(\boldsymbol{p}_0\) and \(\boldsymbol{v}_0\) denote the initial position and initial velocity of the object, \(t_i\) is the time at the \(i\)-th data point, and \(N_p\) is the total number of data points in the trajectory.  
The PDS is defined as the average deviation between the approximated parabolic curve and the actual trajectory:
\begin{align}
\text{PDS} &= \frac{1}{N_p} \sum_{i=1}^{N_p} 
\left\| \boldsymbol{p}_i - \left(\boldsymbol{p}_0 + \boldsymbol{v}_0 t_i + \tfrac{1}{2} \boldsymbol{g} t_i^2 \right) \right\| .
\end{align}
In the evaluation, the PDS was computed for each object as the average over 400 trajectories.

\subsection{Ablation Study on the OAE}
\label{sec:ablation_oae_encoder}

\subsubsection{Setup}
To investigate which model is best suited as the OAE, we compared IE values using the following variants:
\begin{itemize}
    \item \textbf{OIPP-NAE-FC}: A variant of \textbf{OIPP-NAE} where the encoder is replaced with an FC layer.
    \item \textbf{OIPP-NAE-TR}: A variant of \textbf{OIPP-NAE} where the encoder is replaced with a Transformer encoder.
\end{itemize}

\textbf{OIPP-NAE-FC} consists of a single FC encoder, a two-layer LSTM, and a single FC decoder. \textbf{OIPP-NAE-TR} consists of an FC layer followed by a Transformer encoder with positional encoding~\cite{vaswani2017} and a single FC decoder, where the Transformer models temporal dependencies.

\subsubsection{Results}

\begin{figure}[t]
    \centering

    \begin{subfigure}[b]{0.9\linewidth}
        \centering
        \includegraphics[width=\linewidth]{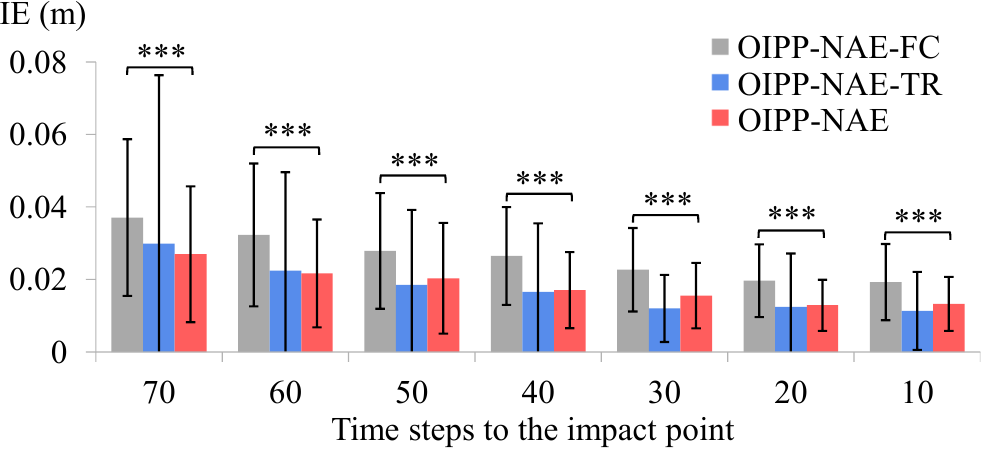}
        \caption{15 seen objects}
        \label{fig:ie_seen_appendix}
    \end{subfigure}

    \vspace{0.5em}  

    \begin{subfigure}[b]{0.9\linewidth}
        \centering
        \includegraphics[width=\linewidth]{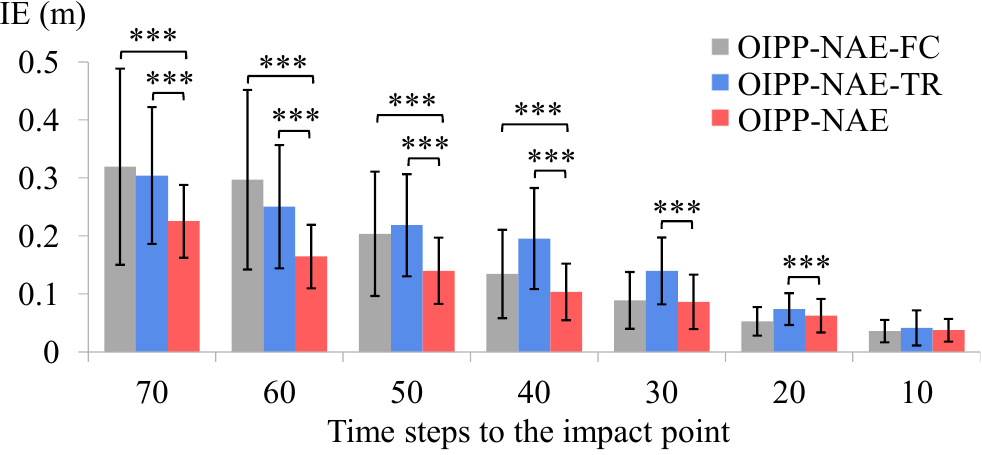}
        \caption{5 unseen objects}
        \label{fig:ie_unseen_appendix}
    \end{subfigure}

    \caption{Comparison of IE with different OAE architectures}
    \label{fig:IE_appendix}
\end{figure}

Fig.~\ref{fig:IE_appendix} shows IE values across different time steps to the impact point. For seen objects, \textbf{OIPP-NAE} outperforms \textbf{OIPP-NAE-FC}, while no significant difference is observed between \textbf{OIPP-NAE} and \textbf{OIPP-NAE-TR}. For unseen objects, \textbf{OIPP-NAE} outperforms both variants. 
We attribute the relatively inferior performance of the Transformer-based encoder to the limited dataset size, which may not fully exploit the its capacity. Further work should clarify how dataset scale affects Transformer encoder performance.

\section*{Acknowledgments}
This work was supported by JSPS KAKENHI Grant Number JP25K21315.






\bibliographystyle{IEEEtran}
\bibliography{IEEEexample}             

\end{document}